\documentclass[conference]{IEEEtran}
\IEEEoverridecommandlockouts
\usepackage{booktabs} % 改善表格样式
\usepackage{multirow} % 合并单元格
\usepackage{array}    % 增强表格功能
\usepackage{adjustbox}% 自动调整内容尺寸
\usepackage{booktabs, adjustbox, makecell, array}
\newcolumntype{C}{>{\centering\arraybackslash}X}  % 居中列类型
\usepackage{cite}
\usepackage{amsmath,amssymb,amsfonts}
\usepackage{algorithmic}
\usepackage{graphicx}
\usepackage{textcomp}
\usepackage{xcolor}
\def\BibTeX{{\rm B\kern-.05em{\sc i\kern-.025em b}\kern-.08em
    T\kern-.1667em\lower.7ex\hbox{E}\kern-.125emX}}
\begin{document}

\title{CRC-SGAD: Conformal Risk Control for Supervised Graph Anomaly Detection\\
% {\footnotesize \textsuperscript{*}Note: Sub-titles are not captured for https://ieeexplore.ieee.org  and
% should not be used}
\thanks{}
}
\bibliographystyle{IEEEtran}

\author{\IEEEauthorblockN{Songran Bai\IEEEauthorrefmark{1}\IEEEauthorrefmark{2},
Xiaolong Zheng\IEEEauthorrefmark{1}\IEEEauthorrefmark{2},
Daniel Dajun Zeng\IEEEauthorrefmark{1}\IEEEauthorrefmark{2}
}
\IEEEauthorblockA{\IEEEauthorrefmark{1}
\textit{State Key Laboratory of Multimodal Artificial Intelligence Systems, Institute of Automation, Chinese Academy of Sciences} \\
\IEEEauthorrefmark{2}\textit{School of Artificial Intelligence, University of Chinese Academy of Sciences} \\
Beijing, China \\
\{baisongran2020, xiaolong.zheng, dajun.zeng\}@ia.ac.cn}
}

\maketitle

\begin{abstract}
Graph Anomaly Detection (GAD) is critical in security-sensitive domains, yet faces reliability challenges: miscalibrated confidence estimation (underconfidence in normal nodes, overconfidence in anomalies), adversarial vulnerability of derived confidence score under structural perturbations, and limited efficacy of conventional calibration methods for sparse anomaly patterns. Thus we propose CRC-SGAD, a framework integrating statistical risk control into GAD via two innovations: (1) A Dual-Threshold Conformal Risk Control mechanism that provides theoretically guaranteed bounds for both False Negative Rate (FNR) and False Positive Rate (FPR) through providing prediction sets; (2) A Subgraph-aware Spectral Graph Neural Calibrator (SSGNC) that optimizes node representations through adaptive spectral filtering while reducing the size of prediction sets via hybrid loss optimization. Experiments on four datasets and five GAD models demonstrate statistically significant improvements in FNR and FPR control and prediction set size. CRC-SGAD establishes a paradigm for statistically rigorous anomaly detection in graph-structured security applications.
\end{abstract}


\section{Introduction}
Graph-based anomaly detection (GAD) has become pivotal in security-critical applications like fraud detection\cite{FinNips2022} and misinformation identification\cite{FakenewsTkdd2024}, leveraging graph neural networks (GNNs) through two paradigms: spatial methods enhance graph homophily through neighborhood redefinition or mitigate over-smoothing via adaptive message passing mechanisms\cite{PcgnnKdd2021, CaregnnCikm2020, H2detWww2022}, while spectral approaches design specialized filters to capture mid- and high-frequency signals induced by anomalous patterns\cite{BwgnnIcml2022, AmnetIjcai2022}.

\begin{figure}[!htbp]
    \centering
    \includegraphics[width=\columnwidth]{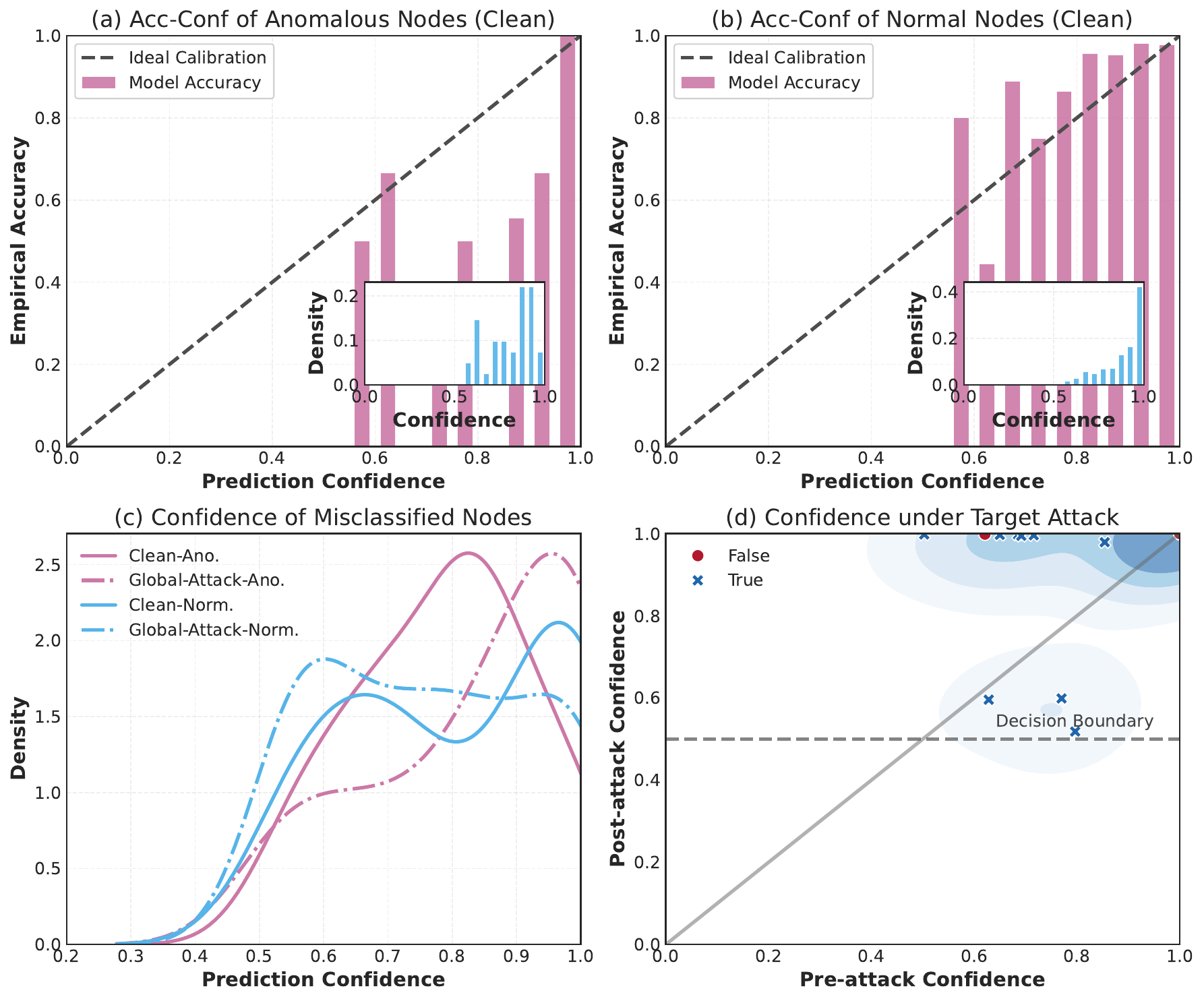}
    \caption{The unreliable prediction confidence of BWGNN on Amazon dataset. (a) and (b) Figures depict the calibration error plots for anomalous nodes and normal nodes, respectively. (c) Figure illustrates the confidence distribution of misclassified nodes (both normal and anomalous) following a global attack. (d) Figure demonstrates the confidence variation of attacked nodes under a targeted attack scenario .}
    \label{fig:fig1}
\end{figure}

However, the deployment of GAD models in high-stakes operational environments faces critical reliability challenges. For instance, malicious entities evade detection by fabricating connections, causing costly misclassifications\cite{AttfakenewsWww2023, AttfraudIjcai2024, AttfakenewsIjcai2024}. Our preliminary experiments reveal three critical insights: (1) Even state-of-the-art models like BWGNN\cite{BwgnnIcml2022} exhibit unreliable confidence scores, displaying overconfidence for anomalous nodes while underestimating certainty for normal nodes (see Fig\ref{fig:fig1}(a)(b)); (2) Subtle targeted structural perturbations\cite{FgaArxiv2019} on correctly classified nodes can induce high-confidence misclassifications (see Fig\ref{fig:fig1}(d)); (3) Global structural perturbations\cite{DiceNhb2018} exacerbate confidence disparity between node classes(see Fig\ref{fig:fig1}(c)). The lack of precise class-dependent uncertainty quantification pose fundamental limitations for reliable GAD models. Traditional calibration methods like Expected Calibration Error (ECE) are unreliable due to insufficient anomalous observations in most confidence bins caused by the inherent rarity. And individual-based calibration method\cite{EiceKdd2023} lacks theoretical guarantees. Conformal Risk Control (CRC)\cite{CrcIclr2024} emerges as a promising alternative through finite-sample statistical guarantees.
We therefore propose the CRC-SGAD framework that generates prediction sets with theoretically guaranteed risk control - the expected FNR and FPR are provably bounded below pre-defined thresholds (e.g., 0.1). Our key contributions are threefold: First, we develop a dual-threshold conformal risk control mechanism tailored for GAD, establishing set-valued functions for normal and anomalous nodes respectively. We prove this adaptation preserves exchangeability assumptions required for valid CRC. Second, to address the observed local heterogeneity in prediction set sizes across neighborhoods, we design a subgraph-aware spectral graph neural calibrator (SSGNC). This component learns frequency-aware node representations through subgraph-specific spectral filters, jointly optimizing prediction inefficiency via a hybrid loss combining conformal risk and weighted cross-entropy. Third, we conduct comprehensive evaluations across four datasets using five classic GAD methods, demonstrating statistically significant improvements in Inefficiency and FNR and FPR control over existing conformal approaches\cite{Conformalbook2023, ApsNips2020, RapsIclr2021}.
% (where central nodes exhibit varying consistency with their neighbors)
\section{Related Work}
\subsection{Graph Anomaly Detection}\label{AA}
GAD methods are typically categorized into supervised (SGAD) and unsupervised (UGAD) paradigms based on label availability\cite{GadTkde2023}. While UGAD primarily employs reconstruction errors or graph contrastive learning for anomaly scoring\cite{AnemoneCikm2021, UgadTKDE2025}, we focus on SGAD due to its superior practicality enabled by partially available trustable labels. Recent advances address heterophily in SGAD through spatial and spectral approaches\cite{GadbenchNips2023}. Spatial methods enhance node representation by redefining neighborhoods or modifying message passing: H2-FDetector\cite{H2detWww2022} differentiates homophilic and heterophilic connections for adaptive message aggregation, PC-GNN\cite{PcgnnKdd2021} employs imbalance-aware subgraph sampling for structural refinement, and PMP\cite{PmpIclr2024} develops a self-tuning mechanism for information fusion from heterogeneous neighbors. Spectral approaches tackle this through adaptive filtering. AMNet\cite{AmnetIjcai2022} captures dual-frequency signals while BWGNN\cite{BwgnnIcml2022} designs beta wavelet-based band-pass filters to preserve anomalous high-frequency patterns.

\subsection{Uncertainty Quantification for GNNs}
Following the taxonomy in deep neural networks, uncertainty in GNNs can also be categorized into aleatoric and epistemic uncertainty\cite{UncertgraphArxiv2024}. Existing approaches primarily quantify uncertainty through maximum softmax probability or entropy\cite{CagcnNips2021}, Bayesian GNNs with Dirichlet priors\cite{GpnNips2021}, frequentist post-hoc methods (e.g., conformal prediction)\cite{CpgnnIcml2023, CfgnnNips2023}, Bayesian stochastic parameter models\cite{BayesiangcnAaai2019}, or ensemble methods\cite{EnsembleTits2024}. Among these, conformal prediction stands out as a model-agnostic frequentist approach that provides theoretical guarantees with minimal additional parameters, high computational efficiency, and robustness. Its extension, conformal risk control\cite{CrcIclr2024}, offers more flexible risk management. For imbalanced graph classification tasks, \cite{EiceKdd2023} extended distribution-based calibration metrics to node-level predictions and proposed Expected Individual Calibration Error (EICE) to address miscalibration issues in rare minority classes. However, while EICE achieves better calibrated probabilities for both normal and anomalous nodes, it remains a point estimation method without theoretical guarantees for controlling false negative and positive rates.

\begin{figure*}[t]
    \centering
    \includegraphics[width=\textwidth]{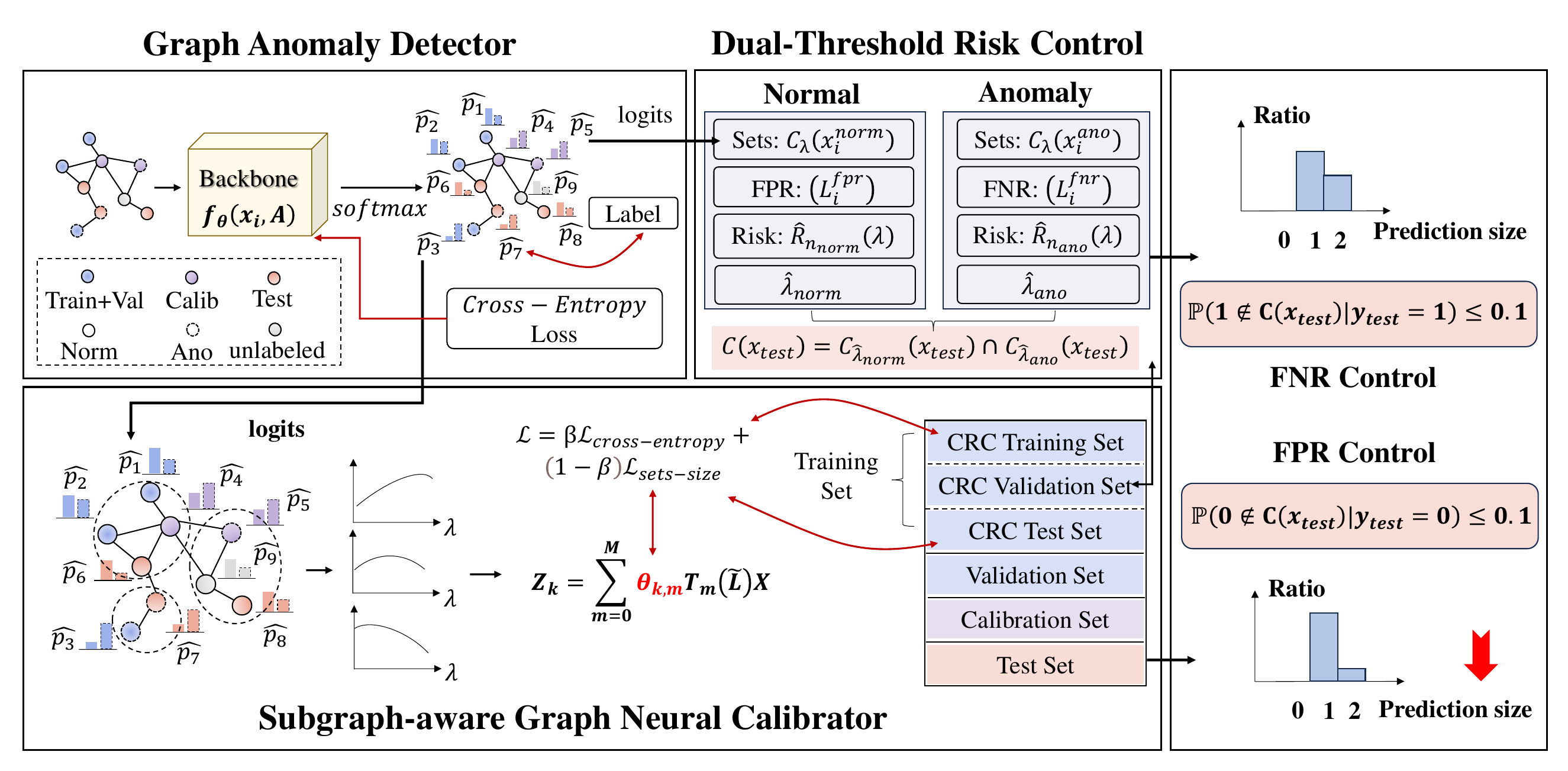}
    \caption{The overall framework of our proposed CRC-SGAD.}
    \label{fig:frame}
\end{figure*}
\section{Preliminaries}
\subsection{Supervised Graph Anomaly Detection}
Consider an undirected graph $G=(V, \mathbf{\Sigma}, \mathbf{A})$ where $V=\{v_i\}_{i=1}^n$ denotes the node set. Each node $v_i \in V$ possesses a feature vector $\mathbf{x}_i \in \mathbb{R}^d$, forming the feature matrix $\mathbf{X} = [\mathbf{x}_1, \mathbf{x}_2, ..., \mathbf{x}_n]^\top \in \mathbb{R}^{n \times d}$. The adjacency matrix $\mathbf{A} \in \{0,1\}^{n \times n}$ encodes edge connections, where $\mathbf{A}_{ij}=1$ iff nodes $v_i$ and $v_j$ are connected. Let $\mathbf{Y} \in \{0,1\}^n$ denote the binary anomaly indicator vector.

The objective is to learn an estimator $f_\theta: \mathbb{R}^d \times \{0,1\}^{n \times n} \rightarrow [0,1]^2$ that predicts anomaly probabilities, where $f_\theta^0(\mathbf{x}_i, \mathbf{A})$ and $f_\theta^1(\mathbf{x}_i, \mathbf{A})$ denote the predicted probabilities for normal and anomalous classes, respectively.
\subsection{Conformal Risk Control}
The conformal risk control framework operates by applying a post-hoc calibration mechanism to refine model $f$'s predictions through a set-valued mapping $C_{\lambda}: \mathcal{X} \to 2^{\mathcal{Y}}$. This mapping generates prediction sets whose conservatism is governed by a threshold parameter $\lambda \in \Lambda$, with larger values of $\lambda$ producing more inclusive sets.

To assess the predictive performance of $C_{\lambda}(\cdot)$, CRC consider a bounded loss function $\ell: 2^{\mathcal{Y}} \times \mathcal{Y} \to (-\infty, B]$ that is non-increasing with respect to $\lambda$. The primary objective is to select an adaptive threshold $\hat{\lambda}$ from calibration data that guarantees:
\begin{equation}
\mathbb{E}_{(x_{n+1}, y_{n+1})}\left[\ell(C_{\hat{\lambda}}(x_{n+1}), y_{n+1})\right] \leq \alpha
\end{equation}
where $\alpha \in (-\infty, B)$ denotes the pre-specified risk tolerance level.

Given $n+1$ exchangeable random functions $\{L_i\}_{i=1}^{n+1}$ where each $L_i: \Lambda \to (-\infty, B]$ is non-increasing and bounded, \cite{CrcIclr2024} define the maximal admissible threshold as:
\begin{equation}
\lambda_{\max} := \sup\left\{ \lambda \in \Lambda \mid L_{n+1}(\lambda) \leq \alpha \right\}
\end{equation}
The calibration procedure utilizes the first $n$ functions to determine $\hat{\lambda}$ that satisfies:
\begin{equation}
\mathbb{E}[L_{n+1}(\hat{\lambda})] \leq \alpha
\end{equation}

The empirical risk estimator is defined as $\hat{R}_n(\lambda) := \frac{1}{n}\sum_{i=1}^{n}L_i(\lambda)$. For desired risk level $\alpha$, compute:
\begin{equation}
\begin{split}
\hat{\lambda} &= \inf \left\{ \lambda \in \Lambda: \frac{n}{n+1}\hat{R}_n(\lambda) + \frac{B}{n+1} \leq \alpha \right\} \\
&= \inf \left\{ \lambda \in \Lambda: \hat{R}_n(\lambda) \leq \alpha - \frac{B-\alpha}{n} \right\}
\end{split}
\end{equation}

The monotonicity of both individual $L_i(\cdot)$ functions and the empirical risk $\hat{R}_n(\cdot)$ permits efficient computation of $\hat{\lambda}$ via binary search with arbitrary precision.

\section{Methodology}
\subsection{Dual-Threshold Conformal Risk Control (DTCRC)}
We adopt the false positive rate (FPR) and the false negative rate (FNR) as our target risk functions. The distributional discrepancy between anomalous and normal nodes in GAD scenarios introduces inherent limitations to single-threshold approaches in simultaneously controlling both FNR and FPR. 
\subsubsection{Risk Functions and Set-Valued Predictors in SGAD}
 For normal samples, the set-valued predictor is defined as:
\begin{equation}
    C_{\lambda_{\text{normal}}}(x_i) = 
    \begin{cases}
        \{1\} & \text{if } f_\theta^1(x_i,A) \geq \lambda_{\text{normal}} \\
        \{0,1\} & \text{otherwise}
    \end{cases}
\end{equation}
with FPR risk $L_i^{\text{FPR}}(\lambda_{\text{normal}}) = \mathbb{P}(0 \notin C_{\lambda_{\text{normal}}}(x_i)|y_i=0) = \mathbb{I}(0 \notin C_{\lambda_{\text{normal}}}(x_i))$. For anomalous samples:
\begin{equation}
    C_{\lambda_{\text{ano}}}(x_i) = 
    \begin{cases}
        \{0\} & \text{if } f_\theta^1(x_i,A) < 1-\lambda_{\text{ano}} \\
        \{0,1\} & \text{otherwise}
    \end{cases}
\end{equation}
with FNR risk $L_i^{\text{FNR}}(\lambda_{\text{ano}}) = \mathbb{P}(1 \notin C_{\lambda_{\text{ano}}}(x_i)|y_i=1) = \mathbb{I}(1 \notin C_{\lambda_{\text{ano}}}(x_i))$. Both risk functions are non-increasing and bounded within [0,1], satisfying the requirements for conformal risk control.

\subsubsection{Exchangeability of Risk Functions}
A sequence $\{(X_i,Y_i)\}_{i=1}^N$ is exchangeable if for any permutation $\sigma$, their joint distribution satisfies $P((X_1,Y_1),\ldots,(X_n,Y_n)) = P((X_{\sigma(1)},Y_{\sigma(1)}),\ldots,(X_{\sigma(n)},Y_{\sigma(n)}))$\cite{Conformalbook2023}. While traditional deep learning assumes independent and identically distributed (i.i.d.) data, SGAD exhibits two distinct properties:
\begin{itemize}
    \item The node-wise random variables exhibit interdependencies with adjacent neighbors, violating the i.i.d. assumption due to their inherent relational dependencies.
    \item The representation learning in GNNs inherently induces structural dependencies through aggregation mechanisms that combine ego-node features with neighborhood information, thereby violating the i.i.d. assumption in their output distributions.
\end{itemize}

Notably, when using permutation-invariant GNN architectures in GAD scenario (e.g., GCN, GAT, GraphSAGE) under random data splits, the model's predictions maintain exchangeability\cite{CfgnnNips2023}. Consequently, the joint distribution of risk functions satisfies:
\begin{equation}
    \mathbb{P}(L_1,\ldots,L_{n+1}) = \mathbb{P}(L_{\sigma(1)},\ldots,L_{\sigma(n+1)})
\end{equation}

\subsubsection{FNR and FPR Control}
Our framework integrates dual-threshold conformal risk control for false positive rate (CRC-FPR) and false negative rate (CRC-FNR), with final predictions derived through their intersection to ensure joint error control. The implementation pipeline comprises four critical stages. First, we train an anomaly detection model $f_\theta: \mathcal{X} \rightarrow \mathbb{R}^{n\times 2}$ through weighted cross-entropy optimization to learn discriminative node representations. During calibration, we strategically partition the data by class membership: For normal nodes, we compute the empirical FPR risk $\hat{R}_{n_{\textnormal{normal}}}(\lambda_{\textnormal{normal}}) = \frac{1}{n_{\textnormal{normal}}}\sum L_i^{\textnormal{FPR}}$ across the calibration set, while for anomalous nodes, we calculate the corresponding FNR risk $\hat{R}_{n_{\textnormal{ano}}}(\lambda_{\textnormal{ano}}) = \frac{1}{n_{\textnormal{ano}}}\sum L_i^{\textnormal{FNR}}$. Subsequently, we employ binary search optimization to determine the optimal thresholds that satisfy the probabilistic error bounds:
\begin{align}
    \hat{\lambda}_{\textnormal{normal}} &= \inf\left\{\lambda : \hat{R}_{n_{\textnormal{normal}}}(\lambda) \leq \alpha - \frac{1-\alpha}{n_{\textnormal{normal}}}\right\}, \\
    \hat{\lambda}_{\textnormal{ano}} &= \inf\left\{\lambda : \hat{R}_{n_{\textnormal{ano}}}(\lambda) \leq \alpha - \frac{1-\alpha}{n_{\textnormal{ano}}}\right\}.
\end{align}
For test instances, the final prediction set $C(x_{\textnormal{test}})$ is obtained through the intersection $C_{\hat{\lambda}_{\textnormal{normal}}}(x_{\textnormal{test}}) \cap C_{\hat{\lambda}_{\textnormal{ano}}}(x_{\textnormal{test}})$, ensuring simultaneous control of both error types. This approach provides provable guarantees that the expected FPR and FNR remain below the predefined significance level $\alpha$.

\begin{table*}[t]
  \caption{Performance Comparison on Amazon and Yelp Datasets}
  \label{tab:results}
  \centering
  \scriptsize  % 缩小字号以适应宽表格
  \begin{adjustbox}{max width=\textwidth}
  \begin{tabular}{@{}l*{12}{c}@{}}  % 去除左右边框，左对齐模型名称
    \toprule
    & \multicolumn{6}{c}{\textbf{Amazon}} & \multicolumn{6}{c}{\textbf{Yelp}} \\
    \cmidrule(lr){2-7} \cmidrule(l){8-13}
    \textbf{Model} & 
    \makecell{Cov} & \makecell{Ine} & \makecell{Amb} & \makecell{Single} & \makecell{FNR} & \makecell{FPR} & 
    \makecell{Cov} & \makecell{Ine} & \makecell{Amb} & \makecell{Single} & \makecell{FNR} & \makecell{FPR} \\
    \midrule
    % \multicolumn{13}{l}{\textit{Graph Convolution Networks}} \\
    \quad GCN        & & & & & & & & & & & & \\
    \quad GCN-TPS    &0.902&1.292&0.292&0.610&0.134&0.094&0.900&1.868&0.868&0.032&0.000&0.116\\
    \quad GCN-APS    &0.999&1.987&0.987&0.012&0.000&0.001&0.999&1.999&0.999&0.000&0.000&0.000\\
    \quad GCN-RAPS   &0.902&1.455&0.478&0.436&0.102&0.084&0.901&1.873&0.873&0.028&0.008&0.115\\
    \quad GCN-Ours   &0.902&1.313&0.313&0.589&\underline{0.098}&\underline{0.098}&0.902&1.693&0.693&0.209&\underline{0.099}&\underline{0.098}\\
    % \addlinespace
    % \multicolumn{13}{l}{\textit{Probabilistic GNN}} \\
    \quad BernNet      & & & & & & & & & & & & \\
    \quad BernNet-TPS  &0.901&1.124&0.124&0.777&0.098&0.099&0.901&1.516&0.516&0.384&0.044&0.109\\
    \quad BernNet-APS  &\textbf{1.000}&1.972&0.972&0.028&0.000&0.000&0.998&1.989&0.989&0.010&0.000&0.002\\
    \quad BernNet-RAPS &0.901&1.250&0.327&0.613&0.069&0.059&0.900&1.596&0.601&0.302&0.048&0.106\\
    \quad BernNet-Ours &0.901&0.935&0.065&\textbf{0.901}&\underline{0.079}&\underline{0.029}&0.901&1.349&0.349&0.553&\underline{0.099}&\underline{0.099}\\
    % \addlinespace
    % \multicolumn{13}{l}{\textit{Attention-based Models}} \\
    \quad AMNet      & & & & & & & & & & & & \\
    \quad AMNet-TPS  &0.901&0.923&0.077&0.901&0.059&0.018&0.900&1.445&0.445&0.455&0.075&0.104\\
    \quad AMNet-APS  &0.901&0.941&0.059&0.901&0.121&0.030&0.999&1.989&0.989&0.010&0.000&0.002\\
    \quad AMNet-RAPS &0.902&1.019&0.166&0.810&0.066&0.020&0.900&1.539&0.547&0.357&0.065&0.101\\
    \quad AMNet-Ours &0.900&0.976&\textbf{0.034}&0.895&\underline{0.078}&\underline{0.070}&0.901&1.369&0.369&0.532&\underline{0.099}&\underline{0.098}\\
    % \addlinespace
    % \multicolumn{13}{l}{\textit{Bandwidth-aware Models}} \\
    \quad BWGNN      & & & & & & & & & & & & \\
    \quad BWGNN-TPS  &0.901&\textbf{0.914}&0.086&0.901&0.047&0.010&0.900&\textbf{1.248}&\textbf{0.248}&\textbf{0.653}&0.147&0.092\\
    \quad BWGNN-APS  &0.936&1.393&0.485&0.497&0.067&0.012&\textbf{0.999}&1.981&0.981&0.018&0.002&0.001\\
    \quad BWGNN-RAPS &0.902&0.976&0.135&0.846&0.065&0.014&0.900&1.360&0.394&0.523&0.124&0.076\\
    \quad BWGNN-Ours &0.900&0.942&0.058&0.900&\underline{0.050}&\underline{0.041}&0.901&1.320&0.320&0.581&\underline{0.100}&\underline{0.098}\\
    % \addlinespace
    % \multicolumn{13}{l}{\textit{Hierarchical Models}} \\
    \quad GHRN       & & & & & & & & & & & & \\
    \quad GHRN-TPS   &0.901&0.918&0.082&0.901&0.043&0.014&0.900&1.335&0.335&0.566&0.096&0.100\\
    \quad GHRN-APS   &0.999&1.975&0.975&0.024&0.006&0.001&0.998&1.984&0.984&0.013&0.003&0.002\\
    \quad GHRN-RAPS  &0.902&0.992&0.147&0.832&0.051&0.017&0.900&1.432&0.454&0.457&0.091&0.089\\
    \quad GHRN-Ours  &0.900&0.932&0.068&0.900&\underline{0.039}&\underline{0.032}&0.901&1.312&0.312&0.590&\underline{0.100}&\underline{0.098}\\
    \bottomrule
  \end{tabular}
  \end{adjustbox}
  
  \vspace{-3mm}  % 微调标题间距
\end{table*}

\subsection{Subgraph-aware Spectral Graph Neural Calibrator}
\begin{figure}[!htbp]
    \centering
    \includegraphics[width=\columnwidth]{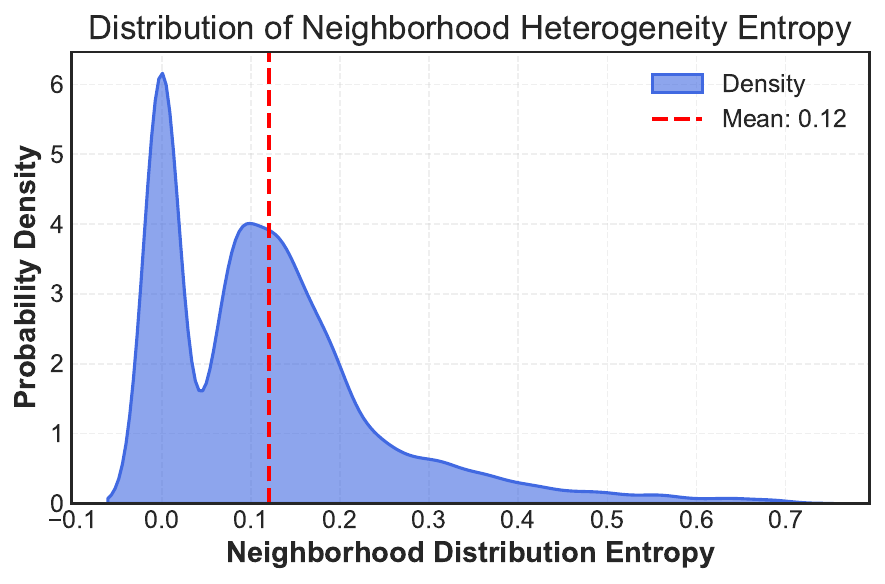}
    \caption{The distribution of Neighborhood Inefficiency entropy of the BWGNN model on the weibo dataset after CRC.}
    \label{fig:fig2}
\end{figure}

Although the framework theoretically maintains both FNR and FPR below $\alpha$, its practical utility is limited by prediction sets containing $\{0,1\}$ or $\emptyset$, which are uninformative outcomes indicating classification uncertainty. To enhance reliability, we prioritize maximizing singleton predictions ({0} or {1}) through calibration-aware learning. Our empirical analysis reveals that the prediction set sizes after DTCRC exhibit subgraph-dependent distributions, corresponding to heterogeneous node-level frequency patterns in graph signals (see Fig.\ref{fig:fig2}). 
Thus we propose Subgraph-aware Spectral Graph Neural Calibrator (SSGNC), featuring: (1) dynamic routing for latent subgraph representation learning, (2) Chebyshev polynomial-based spectral convolution with subgraph-specific filters, and (3) multi-layer architecture with hierarchical feature aggregation. The design explicitly addresses frequency heterogeneity through adaptive spectral filtering across localized subgraph structures.

\subsubsection{Dynamic Routing Mechanism}
Given input node features $H \in \mathbb{R}^{n \times d}$, we learn $K$ latent subgraph representations through an iterative routing process. Let $C \in \mathbb{R}^{K \times d}$ denote the trainable prototype vectors. The routing probabilities are computed as:

\begin{equation}
    s_{ik}^{(0)} = \frac{\exp(\langle h_i, c_k \rangle)}{\sum_{j=1}^K \exp(\langle h_i, c_j \rangle)}
\end{equation}

where $\langle \cdot, \cdot \rangle$ denotes the dot product, $h_i$ is the $i$-th row of $H$, and $c_k$ is the $k$-th prototype. We refine the routing probabilities through $T$ iterations of normalization and prototype adjustment:

\begin{equation}
    c_k^{(t)} = \frac{\sum_{i=1}^N s_{ik}^{(t-1)}h_i}{\sum_{i=1}^N s_{ik}^{(t-1)} + \epsilon}
\end{equation}

\begin{equation}
    s_{ik}^{(t)} = \frac{\exp(\langle h_i, c_k^{(t)} \rangle)}{\sum_{j=1}^K \exp(\langle h_i, c_j^{(t)} \rangle)}
\end{equation}

The final prototype vectors are updated with momentum:
\begin{equation}
    C \leftarrow \beta C + (1-\beta)C^{(T)}
\end{equation}
where $\beta$ is the momentum coefficient set to 0.9 in our implementation.

\subsubsection{Chebyshev Graph Convolution}
For spectral graph filtering, we employ Chebyshev polynomial approximation of graph spectral convolution\cite{ChebyshevNips2022}. Given the normalized graph Laplacian $\tilde{L} = I - D^{-1/2}AD^{-1/2}$, the $m$-th order Chebyshev basis is computed recursively:
\begin{align}
    T_0(\tilde{L})X &= X \\
    T_1(\tilde{L})X &= \tilde{L}X \\
    T_m(\tilde{L})X &= 2\tilde{L}T_{m-1}(\tilde{L})X - T_{m-2}(\tilde{L})X
\end{align}

The subgraph-specific spectral convolution is then formulated as:
\begin{equation}
    Z_k = \sum_{m=0}^M \theta_{k,m}T_m(\tilde{L})X
\end{equation}
where $\theta_{k,m} \in \mathbb{R}^{d \times d}$ are learnable parameters for subgraph $k$ and polynomial order $m$.

\subsubsection{Subgraph-aware Spectral Filtering}
The final node representations are obtained by combining the subgraph-specific convolutions weighted by the routing probabilities:
\begin{equation}
    H^{l+1} = \sigma\left(\sum_{k=1}^K \text{diag}(s_{:,k}^l) Z_k^l\right)
\end{equation}
where $\sigma$ denotes the activation function and $s_{:,k}^l$ represents the routing probabilities for subgraph $k$ at layer $l$.

\subsubsection{Multi-layer Architecture}
The complete SSGNC model with $L$ layers can be formulated as:
\begin{equation}
    H^{(0)} = \text{MLP}_{\text{in}}(X)
\end{equation}
\begin{equation}
    H^{(l)} = \text{Dropout}\left( \sigma\left( \text{SS-Conv}^{(l)}(G, H^{(l-1)}) \right) \right)
\end{equation}
\begin{equation}
    Y = \text{MLP}_{\text{out}}\left( \text{Concat}(H^{(1)}, \dots, H^{(L)}) \right)
\end{equation}
where $\text{SS-Conv}^{(l)}$ denotes our subgraph-aware spectral convolution layer at depth $l$, and $\text{Concat}$ represents feature concatenation.

\section{Experiments}
\subsection{Datasets and Baselines}
\textbf{Datasets}. We utilize four benchmarks spanning social networks and e-commerce: \textit{Reddit}, \textit{Weibo}, \textit{YelpChi}, and \textit{Amazon}. These encompass diverse graph structures including user-subreddit interactions, fraud patterns, and multi-relational reviews. Node features incorporate linguistic attributes (LIWC), text embeddings (FastText), and behavioral metadata.\cite{GadbenchNips2023}.

\textbf{Baseline Models}. We experiment on five GAD models: GCN, BernNet\cite{BernnetNips2021}, AMNet\cite{AmnetIjcai2022}, BWGNN\cite{BwgnnIcml2022}, and GHRN\cite{GhrnWww2023}. For uncertainty quantification, three conformal predictors are evaluated: CP-\textit{TPS} \cite{Conformalbook2023}, CP-\textit{APS} \cite{ApsNips2020}, and CP-\textit{RAPS} \cite{RapsIclr2021}. The variants differ in non-conformity scoring: TPS prioritizes top-class probabilities, APS accumulates softmax values, while RAPS penalizes low-confidence class inclusion.

\subsection{Evaluations}
To assess CRC-SGAD, we adopt six metrics\cite{CpgnnIcml2023}: Coverage (true label inclusion probability), Inefficiency (average set size), and Ambiguity (frequency of indeterminate {0,1} outputs). Singleton Rate tracks single-point predictions, while set-based FPR and FNR quantify error types. These enable rigorous multi-dimensional assessment of statistical reliability and practical utility in uncertainty-aware systems.

\subsection{Main Results}

\subsubsection{Performance Analysis}
Our experiments reveal fundamental limitations in existing conformal prediction approaches. Notably, both the approaches anchored in TPS and RAPS are incapable of ensuring control over the FPR and FNR. While the APS-based method appears to achieve dual error rates below 0.01, this performance comes at the expense of prohibitively large prediction sets. Specifically, average prediction set sizes exceed 1.7 across benchmark datasets. Such inflated uncertainty quantification renders the method impractical for real-world applications.

Conversely, the proposed CRC-SGAD framework more effectively addresses this critical trade-off. The method achieves strict false negative rate control (FNR $<$ 0.01, FPR $<$ 0.01) while maintaining prediction Inefficiency with average set sizes reduced by 15.3-52.8\% compared to APS baselines. 

\subsubsection{Effectiveness of Sub-Modules}
We rigorously validate the effectiveness of our proposed SSGNC framework across five benchmark datasets and five GAD methods. We find the proposed spectral calibration module achieves an average 13.4\% reduction in Singleton ratio across all experimental configurations, while maintaining risk control guarantees. This statistically significant improvement demonstrates the efficacy of our subgraph-aware calibration framework in enhancing the operational efficiency of prediction sets without compromising risk control objectives. 

\begin{table}[t]
  \caption{The improvement of Singleton ratio after SSGNC}
  \label{tab:ablation}
  \centering
  \small  % 适当缩小字号
  \begin{tabular}{@{}ccccc@{}}
    \toprule
    & \textbf{amazon} & \textbf{yelp} & \textbf{reddit} & \textbf{weibo}\\
    \midrule
    GCN    & 9.48\%$\uparrow$ & 45.1\%$\uparrow$ & 1.17\%$\uparrow$ & 2.00\%$\uparrow$\\
    BernNet  & 3.70\%$\uparrow$ & 4.73\%$\uparrow$ & 66.9\%$\uparrow$ & 9.56\%$\uparrow$\\
    AMNet  & 5.80\%$\uparrow$ & 3.70\%$\uparrow$ & 8.31\%$\uparrow$ & 0.22\%$\uparrow$\\
    BWGNN  & 0.00\%$\uparrow$ & 2.29\%$\uparrow$ & 44.9\%$\uparrow$ & 4.43\%$\uparrow$\\
    GHRN   & 0.00\%$\uparrow$ & 3.15\%$\uparrow$ & 52.5\%$\uparrow$ & 0.11\%$\uparrow$\\
    \bottomrule
  \end{tabular}
  \vspace{-2mm}  % 微调标题间距
\end{table}

\subsubsection{Parameter Sensitivity Analysis}
We conduct a systematic evaluation of the BWGNN model on the Yelp dataset to investigate the impact of predefined risk thresholds and the number of SSGNC prototypes on the empirical FPR, FNR, and singleton ratio in prediction sets. As shown in Table \ref{tab:parameter}, our framework demonstrates robust error rate control through adaptive calibration of prediction sets, consistently satisfying the target error rates across all predefined FNR and FPR configurations. The analysis of prototype quantity reveals a non-monotonic relationship with singleton ratio: as the number of prototypes increases from 3 to 7, the singleton ratio first rises (peaking at 5 prototypes) and then declines.

\begin{table}[t]
  \caption{Parameter sensitivity analysis of BWGNN model on Yelp dataset.}
  \label{tab:parameter}
  \centering
  \scriptsize  % 与表2字号统一
  \begin{tabular}{@{}c c c c c c@{}}
    \toprule
    \multicolumn{1}{c}{(FNR, FPR)} & 
    \multicolumn{3}{c}{(0.1, 0.1)} & 
    \multicolumn{1}{c}{(0.05, 0.1)} & 
    \multicolumn{1}{c}{(0.1, 0.05)} \\
    \cmidrule{1-6}
    Prototype Number & 3 & 5 & 7 & 5 & 5 \\
    \cmidrule{1-6}
    FNR       & 0.100 & 0.100 & 0.100 & 0.048 & 0.100 \\
    FPR       & 0.098 & 0.098 & 0.098 & 0.098 & 0.049 \\
    Singleton Ratio & 0.572 & 0.587 & 0.576 & 0.443 & 0.537 \\
    \bottomrule
  \end{tabular}
  \vspace{-2mm}  % 与表2间距统一
\end{table}
\section{Conclusion}
In conclusion, our study addresses the critical need for class-aware uncertainty quantification and calibration in imbalanced graph anomaly detection. Our findings reveal that existing GAD models often yield poorly calibrated confidence estimates, which further deteriorate under structural perturbations. Our CRC-SGAD framework achieves provable statistical guarantees on FNR and FPR while incorporating a subgraph-aware spectral graph neural calibrator that optimizes prediction set sizes without compromising statistical validity. These results collectively highlight the framework's potential to advance reliable graph anomaly detection systems in safety-critical applications.

\bibliography{IEEEexample}
\end{document}